\documentclass[runningheads]{llncs}
\usepackage{hyperref}

\usepackage[T1]{fontenc}
\usepackage{graphicx}
\begin{document}
\title{Efficient Strategy for Improving Large Language Model (LLM) Capabilities}


\author{Julián Camilo Velandia Gutiérrez \inst{}\orcidID{0009-0000-8617-7445}}
\authorrunning{J. Velandia}

\institute{Universidad Nacional de Colombia, Bogotá, Colombia \\
\email{jvelandiag@unal.edu.co} \\
}

\maketitle
%
\begin{abstract}
Large Language Models (LLMs) have become a milestone in the field of artificial intelligence and natural language processing. However, their large-scale deployment remains constrained by the need for significant computational resources. This work proposes starting from a base model to explore and combine data processing and careful data selection techniques, training strategies, and architectural adjustments to improve the efficiency of LLMs in resource-constrained environments and within a delimited knowledge base. The methodological approach included defining criteria for building reliable datasets, conducting controlled experiments with different configurations, and systematically evaluating the resulting variants in terms of capability, versatility, response time, and safety. Finally, comparative tests were conducted to measure the performance of the developed variants and to validate the effectiveness of the proposed strategies. This work is based on the master's thesis in Systems and Computer Engineering titled \textit{Efficient Strategy for Improving the Capabilities of Large Language Models (LLMs)} \cite{1}.

\end{abstract}
\section{Family of models}
This work describes the process through which a family of LLMs is developed, each employing different performance enhancement methods while sharing the same base model \cite{2}. The proposed methodology adopts a quantitative, experimental approach consisting of the following phases.

\subsection{Method Selection}
It is necessary to define how many and which performance enhancement methods will be used, as this determines the size of the model family. To address this, an analysis was conducted to evaluate various approaches for improving LLM performance while taking into account inherent limitations in time, resources, and scope. Priority was given to techniques that are fast, cost-effective, and compatible with other methods.

The approach proposes enhancing the LLM across three general dimensions: quality, response formatting, and efficiency.

\subsubsection{Quality}

Based on the research conducted, Retrieval-Augmented Generation (RAG) was selected as the most suitable technique for improving the quality of the model’s responses. This method enriches outputs by incorporating contextual information retrieved from external databases, which is particularly useful in reducing hallucinations and strengthening prompts with additional data \cite{3}. Although its implementation is more complex and requires longer setup times, a key advantage of RAG is its modular nature, allowing it to be connected to or disconnected from the model’s inference pipeline as needed.

\subsubsection{Response Formatting}

The research identified Fine-Tuning with LoRA \cite{8} as the most appropriate method for enhancing the structure and formatting of the model’s outputs. This approach adapts pre-trained models to specific tasks by adding low-rank, trainable parameters without altering the original weights, making it ideal for customizing the model to meet specific needs. Its primary advantage lies in its ability to produce structured, precise, and consistent outputs aligned with predefined examples. Additionally, Fine-Tuning with LoRA is particularly beneficial in scenarios requiring consistent style, format, or structure in the responses to satisfy specific user or system requirements.

\subsubsection{Efficiency}

Post-Training Quantization \cite{12} was determined to be the most effective technique for improving model efficiency. This method increases the efficiency of pre-trained models by reducing the numerical precision of their weights and activations without requiring a full retraining process, making it well-suited for deployment in environments with limited computational resources. Its main advantage lies in its ability to reduce model size and speed up inference while maintaining acceptable accuracy levels. Furthermore, Post-Training Quantization is especially valuable in scenarios where fast and efficient execution of the model is necessary to meet hardware or user response time requirements.

\subsection{Dataset}

For the selected methods, a knowledge base is required that meets the following criteria:
\begin{itemize}
    \item Contains relatively recent information
    \item Provides accurate and preferably academic content
    \item Includes structured information, ideally addressing complex problem-solving
    \item Is easily accessible and low-cost or free
\end{itemize}

Based on this analysis, the knowledge base selected consisted of doctoral dissertations, master's theses, specialization projects, and undergraduate final papers from the \textit{Universidad Nacional de Colombia}, which are freely available through the UNAL repository \cite{4} (a total of 1,920 documents as of June 2023).

To build a dataset from this information, a scraper was developed to read valid URLs and store them in a temporary dataset with a status flag indicating pending processing. Subsequently, a process was executed to take the pending valid URLs so that another scraper could extract the content of each PDF file and store it in the final dataset, along with the document metadata. This procedure constitutes the knowledge base on which the work and evaluations will be conducted. The extraction process is illustrated in Figure~1 and Figure~2. However, some methods require additional post-processing.
The most relevant data from the dataset are shown in Table~\ref{tab:dataset_statistics}.

\begin{figure}
\includegraphics[width=\textwidth]{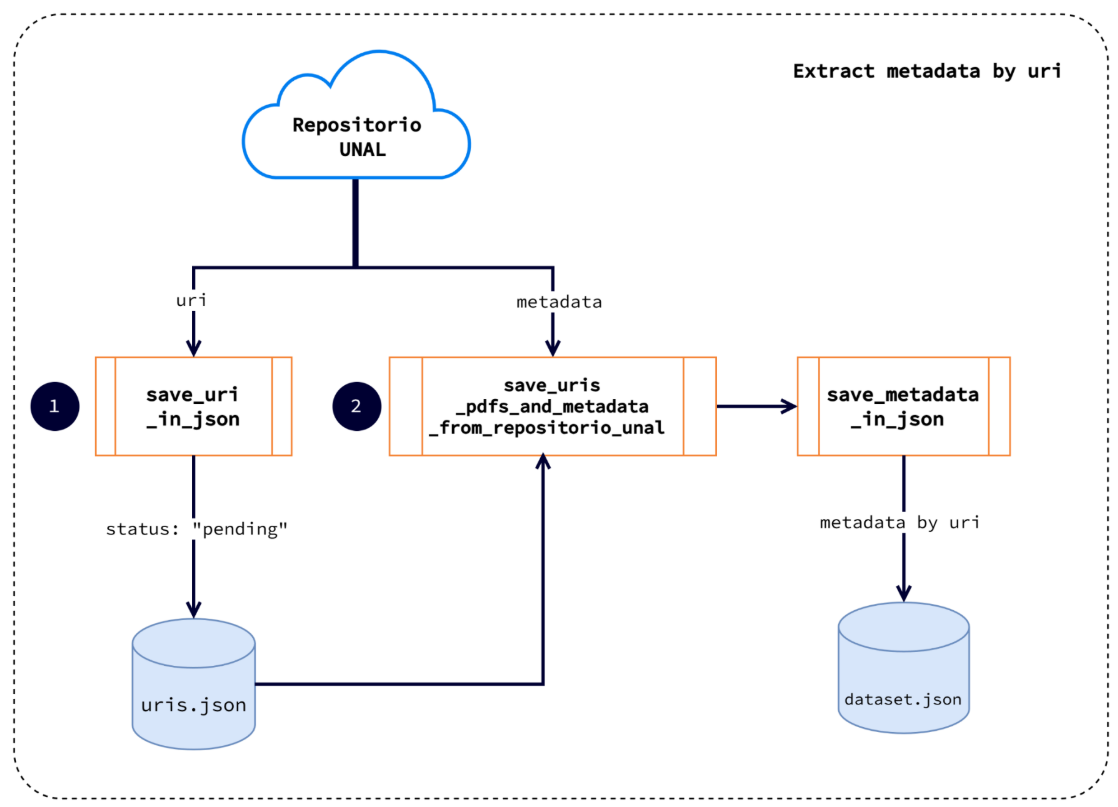}
\caption{Illustration of the URI extraction process from academic theses and dissertations.} \label{fig1}
\end{figure}

\begin{figure}
\includegraphics[width=\textwidth]{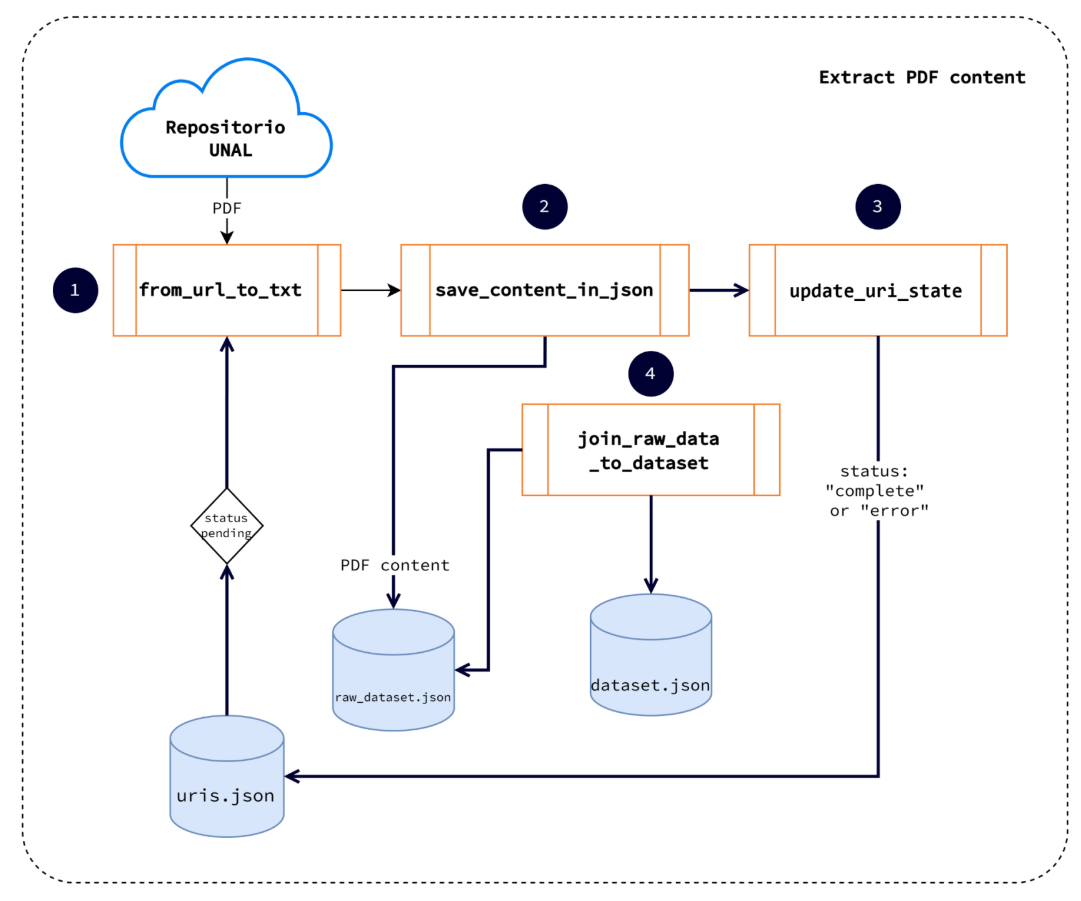}
\caption{Illustration of the content extraction process from each academic thesis.} \label{fig2}
\end{figure}

The final dataset containing the knowledge source used in this work is stored in JSON format as follows:

\begin{samepage}
\begin{verbatim}
{
    "uri": {
        "advisor": "Academic advisor.",
        "author": "Document author.",
        "date": "Publication date.",
        "description": "Document abstract.",
        "title": "Document title.",
        "program": "Academic program.",
        "faculty": "Academic faculty.",
        "raw_content": "Full text content."
    }
}
\end{verbatim}
\end{samepage}

\begin{table}[ht]
\centering
\caption{Dataset Statistics}
\label{tab:dataset_statistics}
\begin{tabular}{ll}
\hline
\textbf{Metric} & \textbf{Value} \\
\hline
Total records & 1910 \\
Extracted texts & 1859 \\
Unique programs & 54 \\
Most frequent program & Medicine - Specialization in Anesthesiology (995 records) \\
Unique advisors & 664 \\
Most frequent advisor & Narváez Rincón, Paulo César (82 records) \\
Unique authors & 1863 \\
Most frequent author & Campos Gaona, Rómulo (4 records) \\
Unique years & 533 \\
Most frequent year & 2014 (182 records) \\
\hline
\end{tabular}
\end{table}

\subsection{Base Model}

The base model refers to the pre-trained model upon which optimization and experimentation are conducted \cite{5}. There is a wide range of open-source models available, and it is ideal to select one with minimal or no prior fine-tuning while remaining under 8B parameters due to computational limitations. At the same time, the model should represent the state of the art among publicly available models. For this work, a 1B parameter version is suitable for optimization tasks. It is also essential to ensure that the model's license allows for free use and modification.

We use LLaMA 3 \cite{6} model was selected for this work due to its recent release, lack of political bias, and sufficient performance for the proposed tasks. Additionally, the widespread adoption and popularity of LLaMA 3 are advantageous when addressing potential issues during implementation \cite{7}.

\subsection{Fine-Tuning with LoRA}

The goal of fine-tuning is to adjust the model's parameters using a new dataset so that it can better adapt to the format and context of this data \cite{8}. However, performing full fine-tuning on LLMs can be highly resource-intensive due to the large number of parameters (1B in this study), making it inefficient and requiring substantial computational power. This challenge is addressed using Low-Rank Adaptation (LoRA), which updates only a small subset of additional trainable parameters rather than all parameters in the model, significantly reducing memory and compute requirements during training \cite{9}.

This process requires the dataset to be structured in a specific way, necessitating additional preprocessing steps.

\subsubsection{Dataset Preparation}

Using the dataset described, a new dataset was created for fine-tuning in a question-answer format, as illustrated in Figure~3.

The content within the rawContent field was split into manageable fragments to ensure coherence and specificity in the generated question-answer pairs. Each fragment was processed with an instruction to generate self-contained question-answer pairs based on the content of the fragment. The resulting dataset was then divided into training and test subsets, following a 75/25 split. These subsets were stored separately to ensure that validation data was not accessible during the training phase \cite{11}.

The average length of prompts was 143 characters, while the average length of responses was 149 characters. The test and validation datasets for fine-tuning were stored in JSON format as shown below:

\begin{samepage}
\begin{verbatim}
[
    {
        "prompt": "According to ISO 15686, how is ... determined?",
        "completion": "It is determined by multiplying the value of ...",
        "fragment": "implemented systems under the standard ..."
    }
]
\end{verbatim}
\end{samepage}

\begin{figure}
\includegraphics[width=\textwidth]{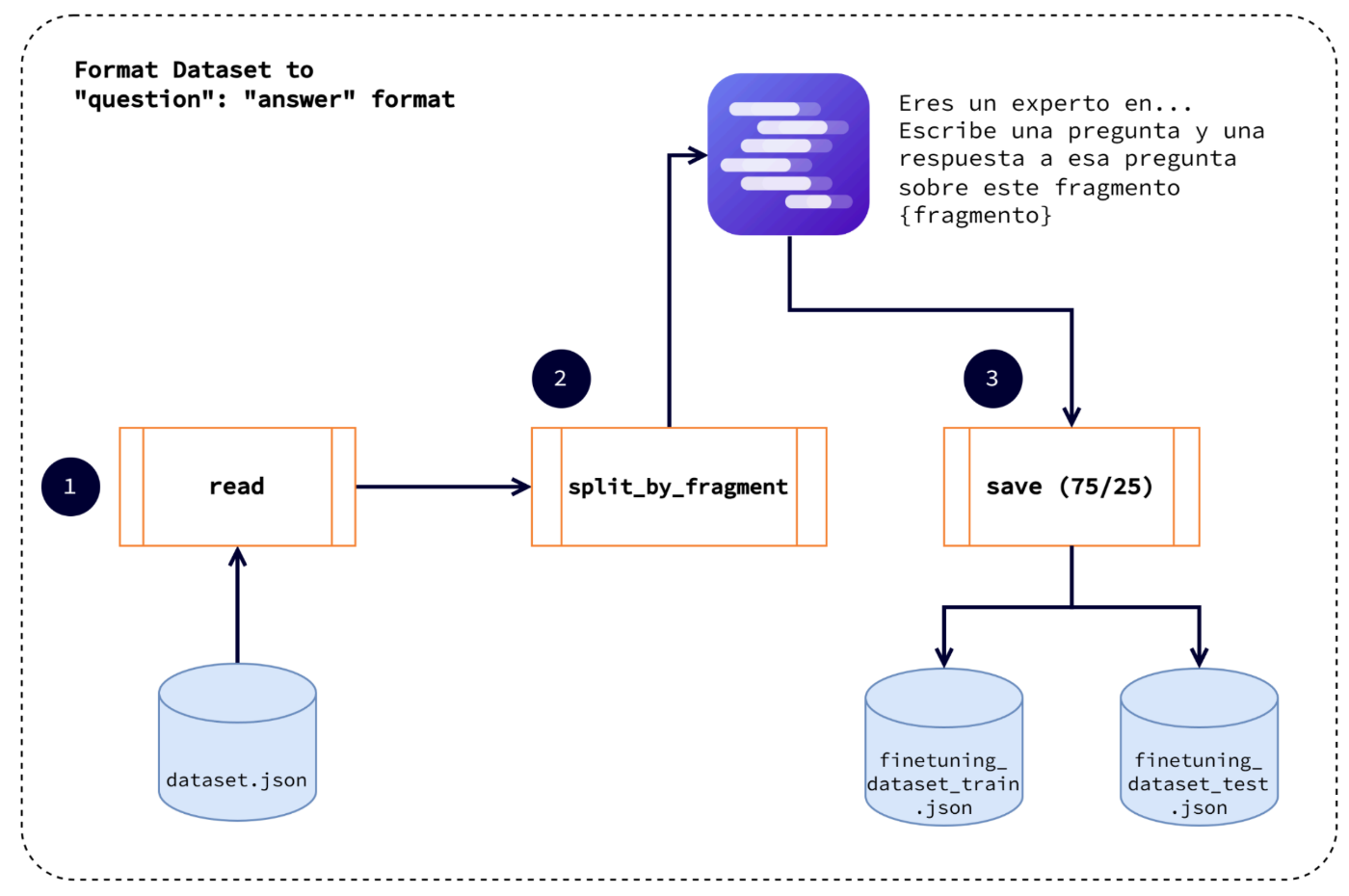}
\caption{Dataset formatting process.} \label{fig3}
\end{figure}

\subsubsection{Train}

For the fine-tuning process, the LLaMA 3 1B base model was used together with the described dataset.

Training was performed using the parameters summarized in Table~\ref{tab:lora_parameters}.

\begin{table}[ht]
\centering
\caption{LoRA training parameters.}
\label{tab:lora_parameters}
\renewcommand{\arraystretch}{1.2}
\begin{tabular}{|p{2.5cm}|p{4.5cm}|p{1.5cm}|p{4.5cm}|}
\hline
\textbf{Parameter} & \textbf{Description} & \textbf{Value} & \textbf{Typical ranges and impact} \\
\hline
LoRA Rank & Dimension of projections in adapted layers. & 2 & Usually 1--8. Higher values allow more adaptation but increase compute needs. \\
\hline
LoRA Alpha & Scaling factor controlling adaptation contribution in the model. & 8 & Usually 1--32. Higher values amplify adaptations but may increase overfitting. \\
\hline
LoRA Dropout & Probability of dropping neurons to prevent overfitting during training. & 0.2 & Usually 0--0.3. Lower values reduce information loss; higher values prevent overfitting but may hinder learning. \\
\hline
\end{tabular}
\end{table}

The base model was lightly quantized to 8 bits. Training was carried out in two configurations, detailed in Table~\ref{tab:quantization_configs}, to evaluate performance differences.

\begin{table}[ht]
\centering
\caption{Quantization configurations during training.}
\label{tab:quantization_configs}
\begin{tabular}{|p{3.5cm}|p{4cm}|p{2cm}|p{4cm}|}
\hline
\textbf{Configuration} & \textbf{Description} & \textbf{Duration} & \textbf{Details} \\
\hline
Base model quantized at 8 bits & LoRA training applied directly to the original model quantized at 8 bits. & ~7 hours & Executed on Google Colab (1 NVIDIA T4 GPU with 16 GB VRAM, 12 GB RAM). \\
\hline
Pre-quantized model at 4 bits & LoRA training applied to the model pre-quantized to 4 bits to evaluate performance differences. & ~7 hours & Executed on Google Colab (1 NVIDIA T4 GPU with 16 GB VRAM, 12 GB RAM). \\
\hline
\end{tabular}
\end{table}

For both configurations, periodic checkpoints were used during training, enabling the process to continue without significant progress loss in case of interruptions or environment limitations. Upon completion, the trained model was uploaded as an open-source checkpoint to the Hugging Face Hub for future use and experimentation.

\subsubsection{Quantization Techniques}

To improve model efficiency, post-training quantization was implemented to reduce the model's size and computational complexity. This approach involves representing the model's parameters with lower precision; in this study, 4-bit quantization was used. This process is illustrated in Figure~4.

\begin{figure}
\includegraphics[width=\textwidth]{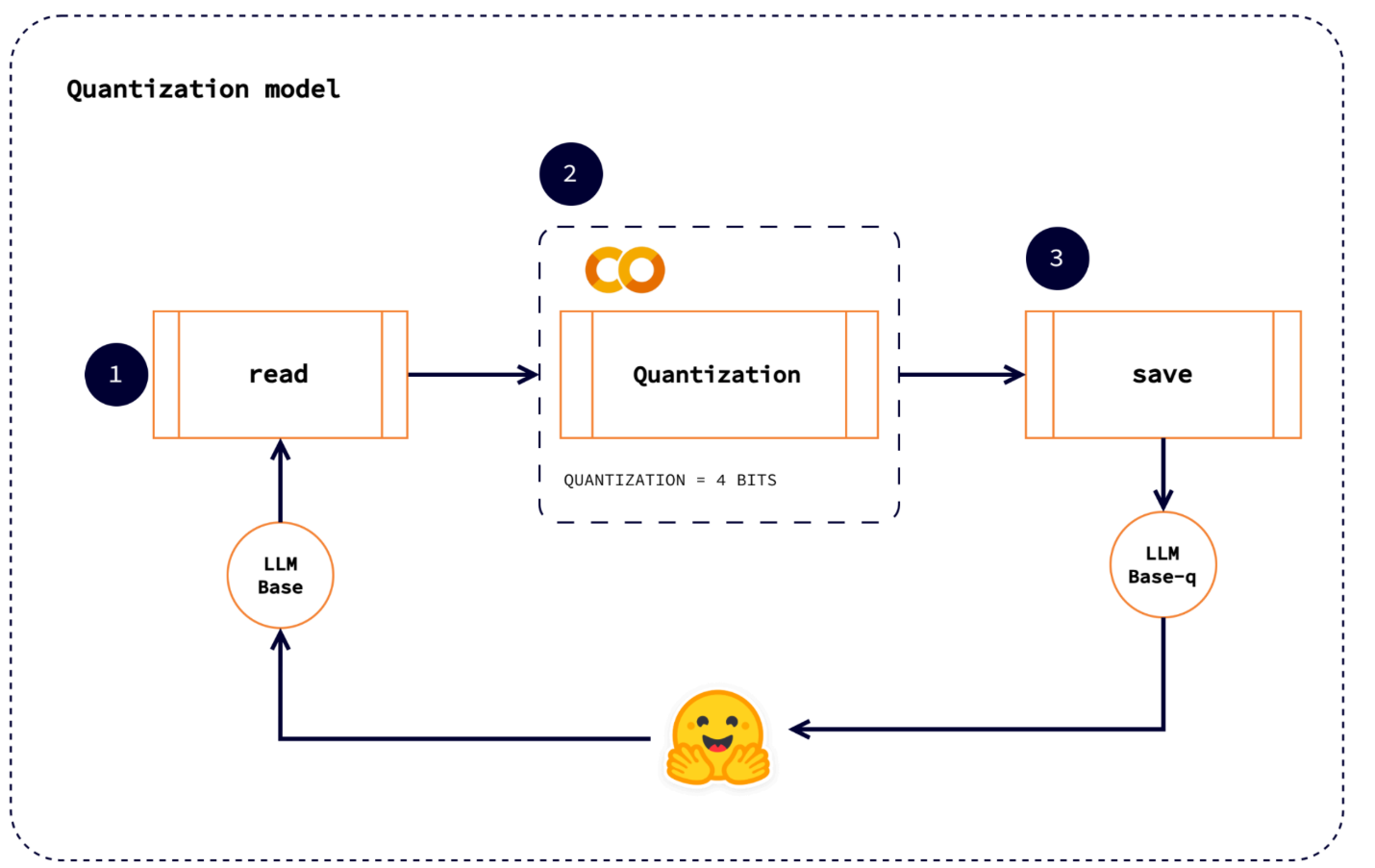}
\caption{Model Quantization Process} \label{fig4}
\end{figure}

Two configurations were evaluated during the quantization process. The first configuration involved quantizing the base model to 4 bits, which took approximately one hour; this quantized version is already available on Hugging Face. The second configuration involved quantizing the fine-tuned model (described in Section 3.1.4) to 4 bits, also taking around one hour, and was executed on Google Colab using a single NVIDIA T4 GPU with 16 GB of VRAM and 12 GB of RAM. Upon completion of the process, the quantized model was uploaded to the Hugging Face Hub as an open-source checkpoint for future use and experimentation.

\subsubsection{Retrieval-Augmented Generation (RAG)}

To enhance the model’s response quality, a Retrieval-Augmented Generation (RAG) system is proposed, structured into two main phases, as illustrated in Figure~3.5.

Preprocessing: In this phase, the dataset is loaded, and each text document is converted into a vector representation using a TF-IDF model \cite{13}. These vectors are stored in binary (.npy) format to ensure efficient retrieval during subsequent queries.

Query: In the query phase, a user input (prompt) is transformed into a vector using the same vectorization model, and its cosine similarity is calculated against the stored dataset vectors. The most relevant documents are selected based on a relevance threshold and a predefined limit on the number of sections retrieved. The initial prompt is then combined with the retrieved texts to create an enriched context that serves as input for downstream tasks.

This approach enables flexible integration of relevant contextual information, optimizing both retrieval and response generation within artificial intelligence systems. Once the system has completed its initial load, queries can be executed almost instantaneously, allowing for efficient use in practical applications.

\begin{figure}
\includegraphics[width=\textwidth]{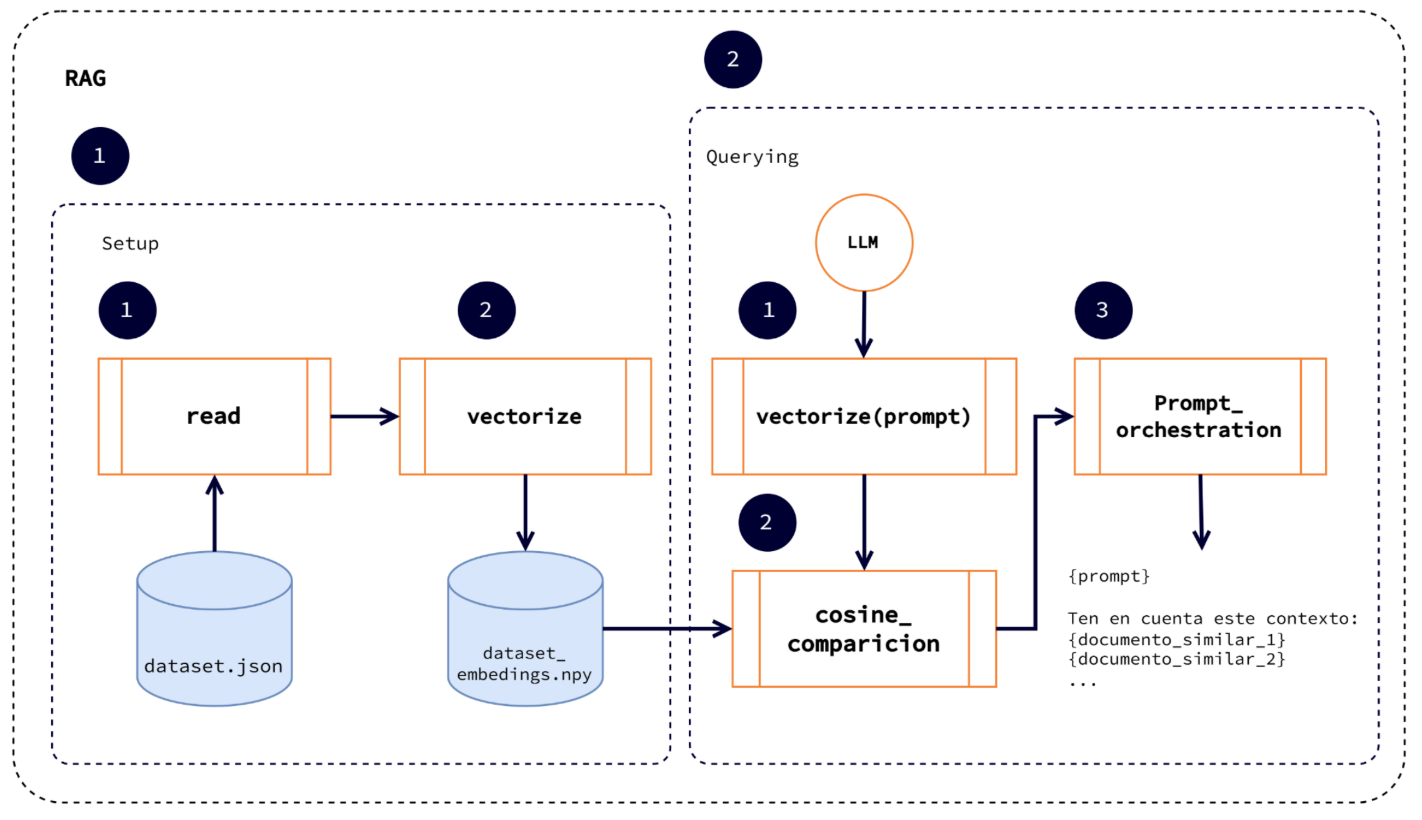}
\caption{Retrieval-Augmented Generation (RAG) System Process} \label{fig5}
\end{figure}

\begin{samepage}
\begin{verbatim}
!pip install SimpleRAGHuggingFace
\end{verbatim}
\end{samepage}

\begin{samepage}
\begin{verbatim}
from rag import Rag
query = "¿Cuál es el Diseño de iluminación, control y
embellecimiento de la cancha del Estadio Alfonso López?"
response = rag.retrieval_augmented_generation(query)
print(response)
"""¿Cuál es el Diseño de iluminación, control y embellecimiento de
la cancha del Estadio Alfonso López?
Keep in mind this context:
Diseño de iluminación ... el Estadio Alfonso López, así como los
resultados obtenidos, entendiendo que un equipo de futbol ..."""
\end{verbatim}
\end{samepage}

The system was developed as a library-based architecture designed to function as an adapter before the query stage and was tested with the models listed in Table~3-7. While the use of this system increased prompt size, it did not exceed the context window of any of the models used. The average execution time for each query was approximately 2000 milliseconds. All tests were conducted on a MacBook Pro M3 equipped with an 8-core performance CPU, a 10-core GPU, and 24 GB of RAM across the following configurations: the base model, the fine-tuned model, the quantized model, the fine-tuned and quantized model, and the quantized fine-tuned model.

\subsubsection{Resulting Models}

At the end of the previous processes, a family of nine variations of the base model was obtained, as illustrated in Figure\~6.

\begin{figure}
\includegraphics[width=\textwidth]{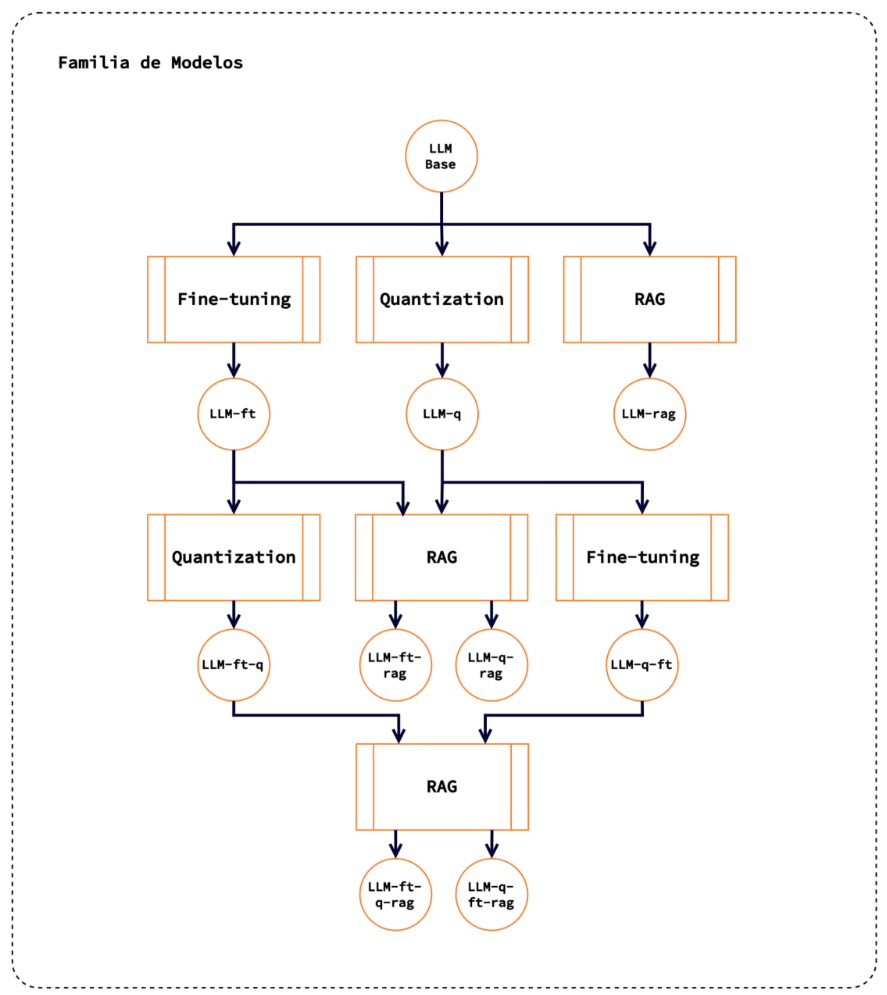}
\caption{Resulting Model Family} \label{fig6}
\end{figure}

The final models were made available on Hugging Face for open access and reproducibility:

\begin{itemize}
    \item \textbf{Base Model:} No training required. 
    
    \texttt{lmstudio-community/Llama-3.2-1B-Instruct-GGUF/Llama-3.2-1B-Instruct-Q8\_0.gguf} (+RAG variant available).
    \item \textbf{Quantized Model:} No training required. 
    
    \texttt{lmstudio-community/Llama-3.2-1B-Instruct-GGUF/Llama-3.2-1B-Instruct-Q4\_K\_M.gguf} (+RAG variant available).
    \item \textbf{Fine-Tuned Model:} ~7 hours of training. 
    
    \texttt{JulianVelandia/Llama-3.2-1B-unal-instruct-ft-gguf/model-f16.gguf} (+RAG variant available).
    \item \textbf{Fine-Tuned Quantized Model:} ~1 hour of training. 
    
    \texttt{JulianVelandia/Llama-3.2-1B-unal-instruct-ft-gguf/model-q4\_k\_m.gguf} (+RAG variant available).
    \item \textbf{Quantized with Fine-Tuning Model:} ~7 hours of training. 
    
    \texttt{JulianVelandia/Llama-3.2-1B-unal-instruct-q-ft-gguf/model-f16.gguf} (+RAG variant available).
\end{itemize}

All RAG configurations did not require additional training and allow for context-augmented retrieval during inference.

\section{Evaluation of the LLMs}

Evaluating the performance of LLMs is particularly challenging because, unlike traditional machine learning models, it is not feasible to calculate an exact error rate. In conventional machine learning workflows, the data sets are divided into training, validation, and test sets, with the model adjusting its parameters during training to minimize a loss function. The model’s ability to generalize is then assessed using the validation and test sets, providing an objective and relatively straightforward method to evaluate performance.

However, LLMs present an additional complexity: for a single prompt, there may be multiple acceptable responses, making exact token-level comparisons unreliable. An alternative to address this challenge is to use an LLM-based evaluation, where a more advanced or specialized model reviews and scores the generated output. This approach enables the assessment of coherence, relevance, and precision without relying solely on human reviewers. Techniques such as LLM-as-a-judge \cite{10} can be applied, allowing a model to compare generated answers against reference responses and assign a score based on predefined evaluation criteria.

\subsection{LLM-Assisted Evaluation}

It is proposed to generate a ranking for a series of questions using a larger, publicly available LLM to perform the task of LLM-as-a-judge, as illustrated in Figure~7.

\begin{figure}
\includegraphics[width=\textwidth]{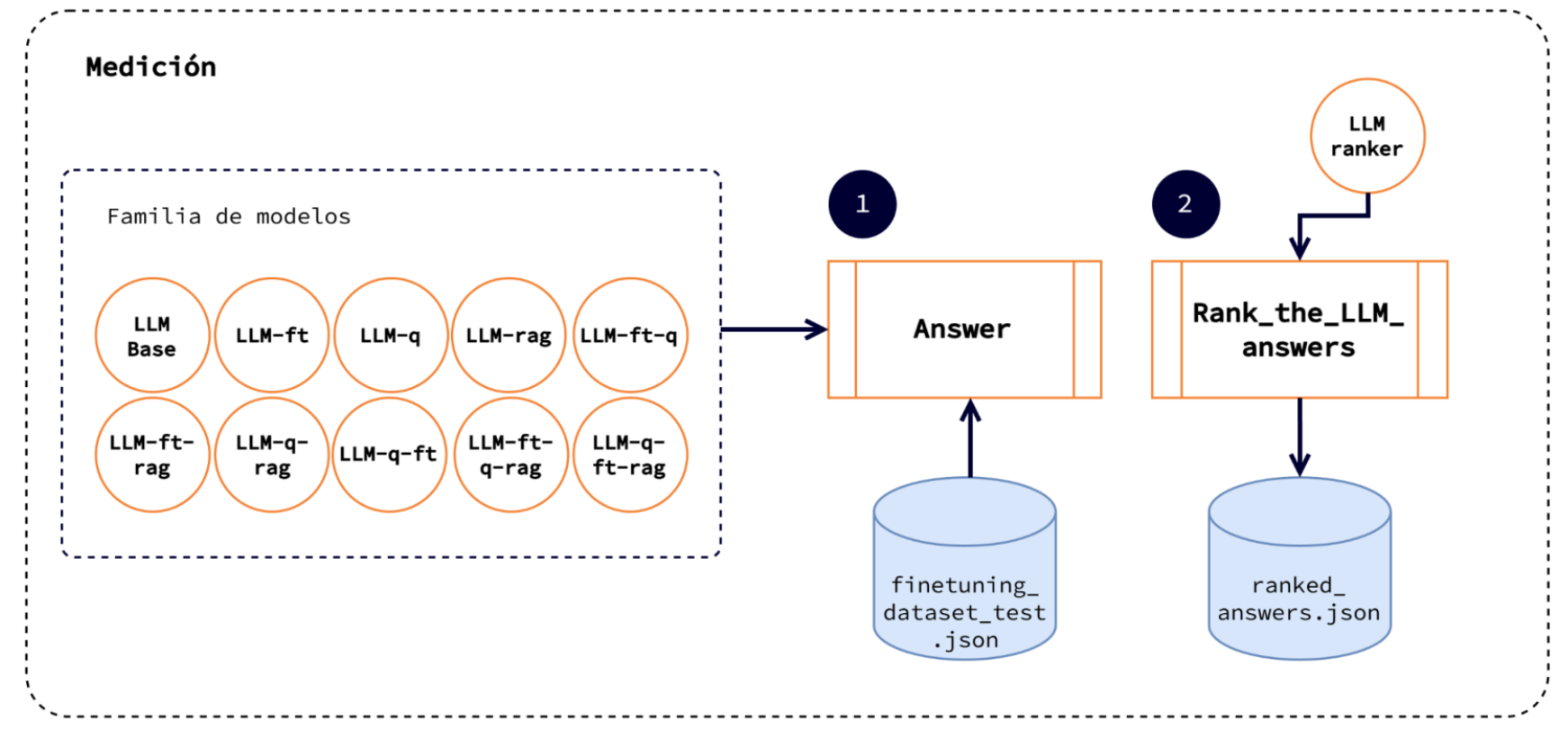}
\caption{Measurement Methodology Using the LLM-as-a-Judge Paradigm} \label{fig7}
\end{figure}

For this evaluation, the GPT-4o model from OpenAI was used with an LLM-as-a-judge prompt. A total of 100 questions were presented to each model, and the ranking process was executed accordingly. The results are presented in Table~9.

\begin{table}[ht]
\centering
\caption{Evaluation results of the implemented models.}
\label{tab:model_evaluation_results}
\renewcommand{\arraystretch}{1.2}
\begin{tabular}{|p{3cm}|p{2.5cm}|p{2.8cm}|p{6cm}|}
\hline
\textbf{Model} & \textbf{Position} & \textbf{First Places} & \textbf{Description} \\
\hline
LLM-q-ft-rag & 2.50 & 26 & Quantized model with fine-tuning and RAG, the best overall performance. \\
\hline
LLM-ft-rag & 2.69 & 22 & Fine-tuned model with RAG, performance close to the best. \\
\hline
LLM-q-ft & 2.89 & 27 & Quantized model with fine-tuning, most frequently ranked first. \\
\hline
LLM-ft & 3.02 & 18 & Fine-tuned model with good performance. \\
\hline
LLM-q & 5.84 & 4 & Quantized model with intermediate performance. \\
\hline
LLM-q-rag & 5.95 & 2 & Quantized model with RAG, rarely in first place. \\
\hline
LLM & 7.47 & 0 & Base model with lower performance. \\
\hline
LLM-rag & 7.63 & 1 & Base model with RAG, without significant improvements. \\
\hline
LLM-ft-q-rag & 8.36 & 0 & Fine-tuned quantized model with RAG, low performance. \\
\hline
LLM-ft-q & 8.56 & 0 & Fine-tuned quantized model, lowest performance. \\
\hline
\end{tabular}
\end{table}

\begin{figure}
\includegraphics[width=\textwidth]{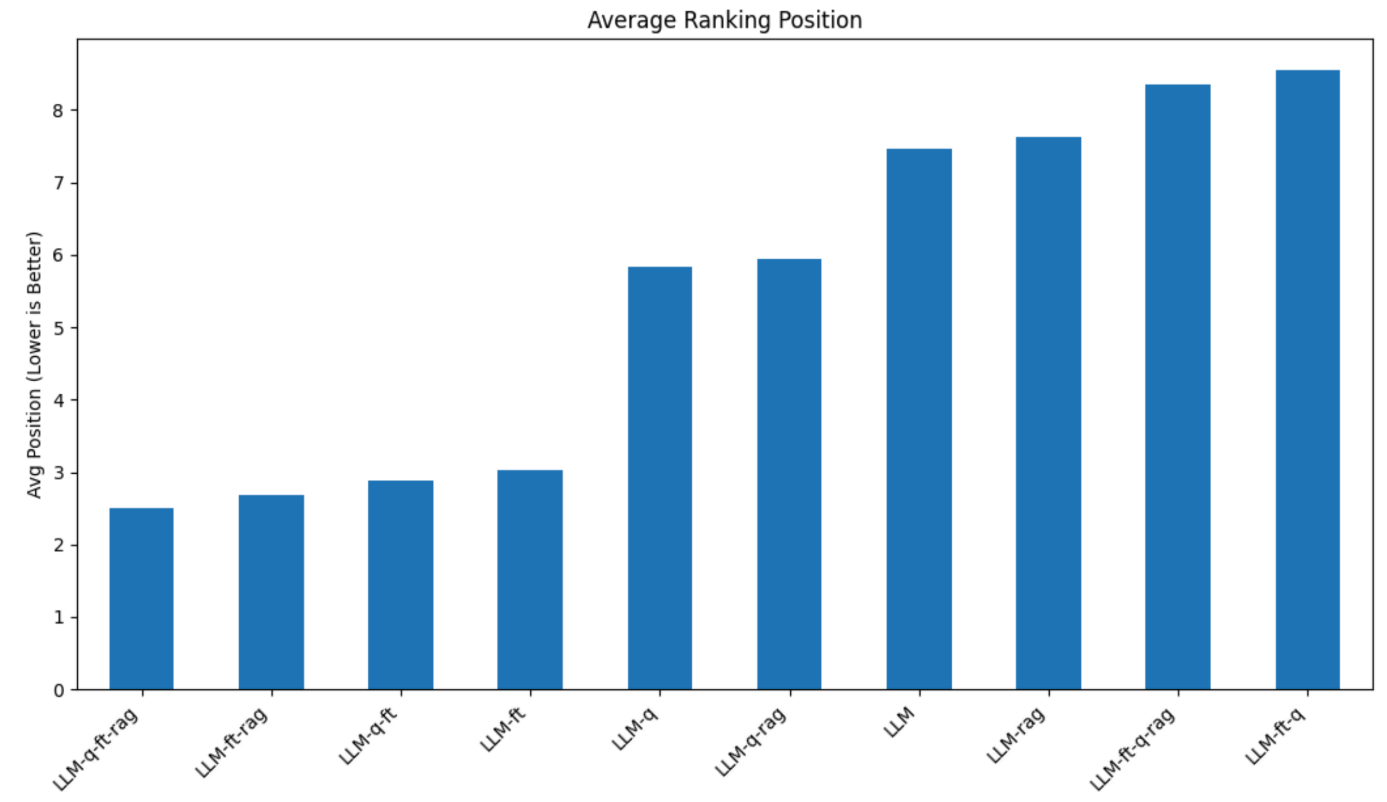}
\caption{Average Position of Each Model} \label{fig8}
\end{figure}

\begin{figure}
\includegraphics[width=\textwidth]{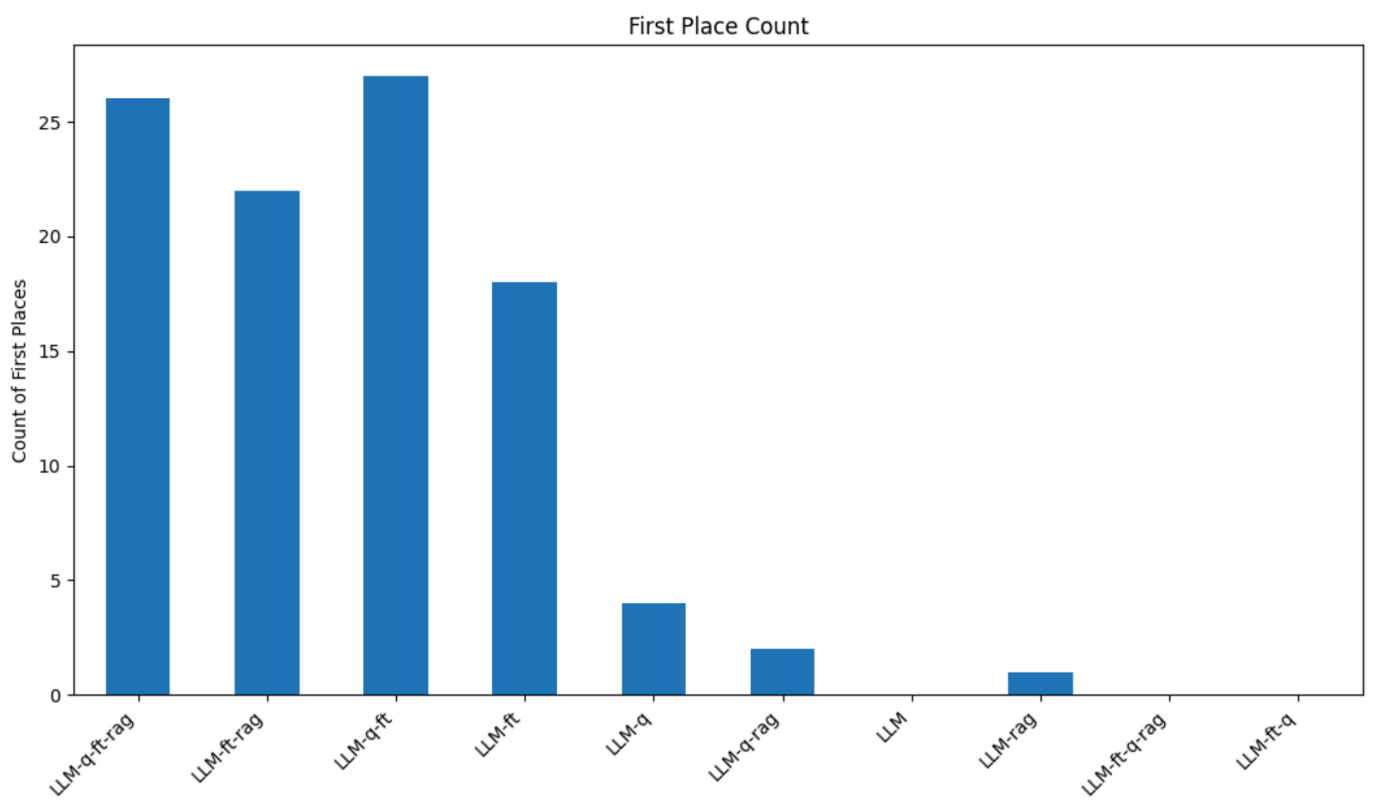}
\caption{Distribution of First-Place Rankings} \label{fig9}
\end{figure}

The results show that the best-performing model is LLM-q-ft-rag, with an average position of 2.50 and ranking first 26 times. However, LLM-q-ft achieved the highest number of first-place rankings (27) despite having a slightly higher average position of 2.89. Overall, models with fine-tuning outperform the base models, while the incorporation of RAG slightly improves performance.

It is also notable that quantization applied to an already fine-tuned model tends to degrade its performance. This occurs because quantization reduces the numerical precision across the entire model, including the adapter trained during fine-tuning, leading to a drop in performance.

The best approach is to first quantize the base model and then apply fine-tuning with LoRA, as this allows the adapter to retain full precision, preserving the benefits of fine-tuning without performance degradation. Additionally, connecting a RAG system provides a slight further improvement in performance.

\section{Conclusions and Recommendations}

\subsection{Conclusions}

Based on the work carried out, it can be concluded that the objectives set at the beginning of the project were achieved. A family of models with different configurations was successfully built, and a benchmarking process was conducted to identify which configurations offered the best relative performance. However, it is important to note that the scope of the project was limited to a specific dataset, employing particular techniques on a single base model. For future iterations, it is recommended to expand the experimental scope to include multiple datasets, different base models, and various combinations of techniques to obtain a deeper and more generalizable understanding of the strategies that enhance LLM performance.

The results presented in Table~4 reveal expected performance patterns among the evaluated models. Overall, models incorporating Retrieval-Augmented Generation (RAG) demonstrate slightly better results, suggesting that access to external information is a key factor in improving the generation of relevant content.

The results demonstrate that fine-tuning significantly improves model performance compared to the base version, achieving lower average rankings and more first-place positions. However, applying quantization after fine-tuning with LoRA led to a noticeable degradation in performance, as seen with models like LLM-ft-q, which ranked among the lowest and did not achieve any first-place results.

On the other hand, quantized models performed similarly or even better than their non-quantized counterparts. They tend to generate shorter and more concise responses while offering substantial improvements in memory consumption and inference speed, making them a viable alternative when computational efficiency is a priority. These findings suggest that using RAG in combination with quantization offers the best balance between response quality and resource optimization. Although fine-tuning provides clear quality improvements, it is not always the most practical option when efficiency constraints are considered.

Based on the work conducted, key takeaways for optimizing models under different priorities are as follows: quantization (Q) enables fast inference with low memory usage at no additional cost; fine-tuning (FT) enhances output quality if resources allow; combining quantization with fine-tuning (Q-FT) allows models to retain precision while improving efficiency; fine-tuning with RAG (FT-RAG) achieves the highest response quality but requires significant computational resources; and combining quantization, fine-tuning, and RAG (Q-FT-RAG) strikes a balance between quality and efficiency, despite RAG’s computational demands.

It is also important to note that the computational setup and the procedures documented in the associated repositories were refined through an iterative process of trial and error, aiming to find the best possible balance between efficiency and availability. A significant challenge encountered during the project was converting LoRA fine-tuned models to the GGUF format, as it was insufficient to convert only the adapter. It was necessary to download the full base model weights, merge them with the fine-tuning weights, and then convert the combined model, which required substantial RAM and often caused runtime crashes. As a result, the conversion process had to be adapted specifically for each model to complete successfully.

\subsection{Recommendations}

To ensure the reproducibility of this work and to facilitate the implementation of the methods used in evaluating language models, a series of strategies are recommended to optimize both computational efficiency and the organization of resources.

It is advisable to use platforms that support GPU or TPU acceleration, such as Google Colab Pro or local environments with optimized hardware, particularly for fine-tuning tasks that require intensive memory and processing capabilities. Additionally, it is recommended to leverage open repositories like the Hugging Face Hub to store and share the resulting models, ensuring accessibility and enabling comparisons with future implementations.

The SimpleRAGHuggingFace library developed for this project facilitates the creation and deployment of RAG systems, provided that the dataset is available on Hugging Face. Its use is recommended for practical applications due to its effectiveness in improving response accuracy and relevance. However, it is essential to manage the knowledge base carefully, ensuring that the data is updated regularly and sourced from reliable repositories to minimize bias in model outputs.

\subsection{Future Work}

The field of artificial intelligence is experiencing unprecedented growth, with state-of-the-art advancements evolving at an accelerated pace, making it challenging to keep up. During the development of this work, various optimization techniques for large language models (LLMs) have been published by leading companies and renowned researchers. In this context, recent studies complement the findings presented in this research, with the following areas standing out:

\subsubsection{Reasoning Paradigm}

Recent research on the reasoning paradigm in LLMs aims to enhance their capabilities in complex tasks using strategies such as Chain-of-Thought (CoT) \cite{14}, which allows problems to be broken down into verifiable intermediate steps. These approaches have significantly improved model performance in reasoning tasks (System 2), such as mathematical problem-solving. Additionally, integrating formal languages like Python and incorporating search algorithms and reinforcement learning have optimized the generation and validation of responses.

A key aspect of these reasoning models, similar to our approach, is the necessity of high-quality, curated training data, which we addressed through the preprocessing of our dataset. While reinforcement learning is the predominant method for training reasoning models, fine-tuning also remains a valid and effective strategy. Although our study did not directly explore prompt structuring or step-by-step verification, the findings suggest that combining information retrieval with self-evaluation strategies could enhance the generation of precise, well-founded responses in future developments.

\subsubsection{DeepSeek-R1 and Its Contribution to Reasoning in LLMs}

The AI research group DeepSeek released the DeepSeek-R1-Zero and DeepSeek-R1 models \cite{15}, designed to enhance reasoning in LLMs through reinforcement learning (RL). DeepSeek-R1-Zero was trained exclusively using RL without prior supervised fine-tuning, resulting in emergent reasoning behaviors but also issues such as low readability and mixed-language outputs. To address these deficiencies, DeepSeek-R1 applies a multi-stage training process with cold-start data before RL, achieving significant improvements. The study also explores model distillation into smaller versions while maintaining competitive performance in reasoning tasks, positioning DeepSeek-R1 as a state-of-the-art open-source model.

In comparison with our work, we share the goal of optimizing models to enhance reasoning capabilities, exploring approaches such as RAG and quantization to achieve efficiency without sacrificing accuracy. However, DeepSeek-R1 prioritizes RL over supervised fine-tuning or RAG, reflecting a trend in the field toward more autonomous learning methods. While our work highlights the effectiveness of RAG in information retrieval to improve responses, DeepSeek-R1 focuses on intrinsic model improvement through RL and distillation.

\subsubsection{Repositories}

The work carried out in this project was published as open-source code to contribute to the community and ensure transparency and reproducibility of the pipeline. The extraction of the UNAL thesis repository dataset was openly published to enable public access and facilitate the training of language models with local academic content, using a structured scraper workflow and made available by Velandia \cite{16}. Subsequently, the dataset was processed and cleaned to support instruction-based tasks and retrieval-augmented generation experiments, following open practices for dataset structuring and also published by Velandia \cite{17}. To enable efficient RAG workflows, the \texttt{SimpleRAGHuggingFace} library was developed and shared, providing modular tools for fast context retrieval within Hugging Face pipelines, made openly available for academic and practical experimentation by Velandia \cite{18}. For model optimization, the quantization of the LLaMA-3.2-1B-Instruct base model was conducted to facilitate its deployment on low-resource hardware while preserving performance, with the quantized model shared by Velandia \cite{19}. An efficient fine-tuning strategy was then implemented on the base model using the prepared dataset to enhance its local academic domain capabilities, following open and reproducible workflows and made available by Velandia \cite{20}. The resulting fine-tuned model was shared for community exploration and integration into academic pipelines \cite{21}, and its quantized version was released to ensure usability across devices with different computational capacities \cite{22}. Further, fine-tuning was performed on a pre-quantized version of the model to explore advanced specialization and compression strategies while preserving accuracy, and this model was released publicly by Velandia \cite{23}. The resulting fine-tuned quantized model was also shared to support comparative benchmarking in diverse environments \cite{24}. Finally, a comprehensive benchmarking suite was developed and published to systematically evaluate the different model versions, quantization levels, and pipeline configurations, promoting open evaluation practices in the local LLM ecosystem, with the repository made publicly available by Velandia \cite{25}.

%
%
%
%

\end{document}